\documentclass{article}

\usepackage[nonatbib, preprint]{neurips_2022}

\usepackage[utf8]{inputenc} % allow utf-8 input
\usepackage[T1]{fontenc}    % use 8-bit T1 fonts
\usepackage{hyperref}       % hyperlinks
\usepackage{url}            % simple URL typesetting
\usepackage{booktabs}       % professional-quality tables
\usepackage{amsfonts}       % blackboard math symbols
\usepackage{nicefrac}       % compact symbols for 1/2, etc.
\usepackage{microtype}      % microtypography
\usepackage{xcolor}         % colors
\usepackage{comment}
\usepackage{xspace}

\newcommand{\NORMAN}[1]{{\bf\textcolor{green}{Norman: #1}}}
\newcommand{\DOMINIK}[1]{{\bf\textcolor{cyan}{Dominik: #1}}}

\newcommand{\DATASET}{MOTFront\xspace}
\title{3D Multi-Object Tracking with \\ Differentiable Pose Estimation}

\author{%
  Dominik Schmauser \\
  Technical University of Munich\\
  \texttt{dominik.schmauser@tum.de} \\
  \And
  Zeju Qiu \\
  Technical University of Munich\\
  \texttt{zeju.qiu@tum.de} \\
  \And
  Norman Müller \\
  Technical University of Munich\\
  \texttt{norman.mueller@tum.de} \\
  \And
  Matthias Nießner \\
  Technical University of Munich\\
  \texttt{niessner@tum.de} \\
}

\begin{document}

\maketitle

\begin{abstract}

We propose a novel approach for joint 3D multi-object tracking and reconstruction from RGB-D sequences in indoor environments.
To this end, we detect and reconstruct objects in each frame while predicting dense correspondences mappings into a normalized object space. We leverage those correspondences to inform a graph neural network to solve for the optimal, temporally-consistent 7-DoF pose trajectories of all objects.
The novelty of our method is two-fold:
first, we propose a new graph-based approach for differentiable pose estimation over time to learn optimal pose trajectories;
second, we present a joint formulation of reconstruction and pose estimation along the time axis for robust and geometrically consistent multi-object tracking.
In order to validate our approach, we introduce a new synthetic dataset comprising 2381 unique indoor sequences with a total of 60k rendered RGB-D images for multi-object tracking with moving objects and camera positions derived from the synthetic 3D-FRONT dataset.
We demonstrate that our method improves the accumulated MOTA score for all test sequences by 24.8\% over existing state-of-the-art methods. In several ablations on synthetic and real-world sequences, we show that our graph-based, fully end-to-end-learnable approach yields a significant boost in tracking performance.
\end{abstract}

\section{Introduction\label{intro}}
Multi-object tracking (MOT) is a key component in many applications such as robot navigation, autonomous driving, or mixed reality. In the outdoor setting, we see significant progress, particularly in the context of LiDAR-based object tracking. In the indoor setting, however, reliable multi-object tracking remains in its infancy. Here, we naturally observe a high level of occlusion, large inter-class variety, and strong appearance changes that severely hamper tracking performance. In addition, we notice that in contrast to the autonomous driving or pedestrian tracking scenarios where large annotated tracking datasets exist, there is no equivalent available for indoor environments.

In the indoor setting, prior works hence often tackle this task by relying on strong 2D/3D detectors followed by an uncoupled data association step.
For object matching, several frame-to-frame heuristics or learned-similarity or geometry-based approaches have been proposed. However, as those modules do not inform each other, this often leads to sub-optimal tracking performance. 

We introduce a holistic approach for joint pose estimation, 3D reconstruction, and data association over time for reliable object pose tracking to address these shortcomings.
We leverage differentiable pose estimation together with a graph neural network for object association in order to obtain temporally consistent 7-DoF object poses (3 rotations, 3 translations and 1 scale). 
By jointly learning to estimate object shapes, we obtain additional feature priors that help to facilitate the association of rigidly-moving objects over time.

\begin{figure}[t!]
  \centering
  \includegraphics[width=1\textwidth, trim=61 425 154 268, clip]{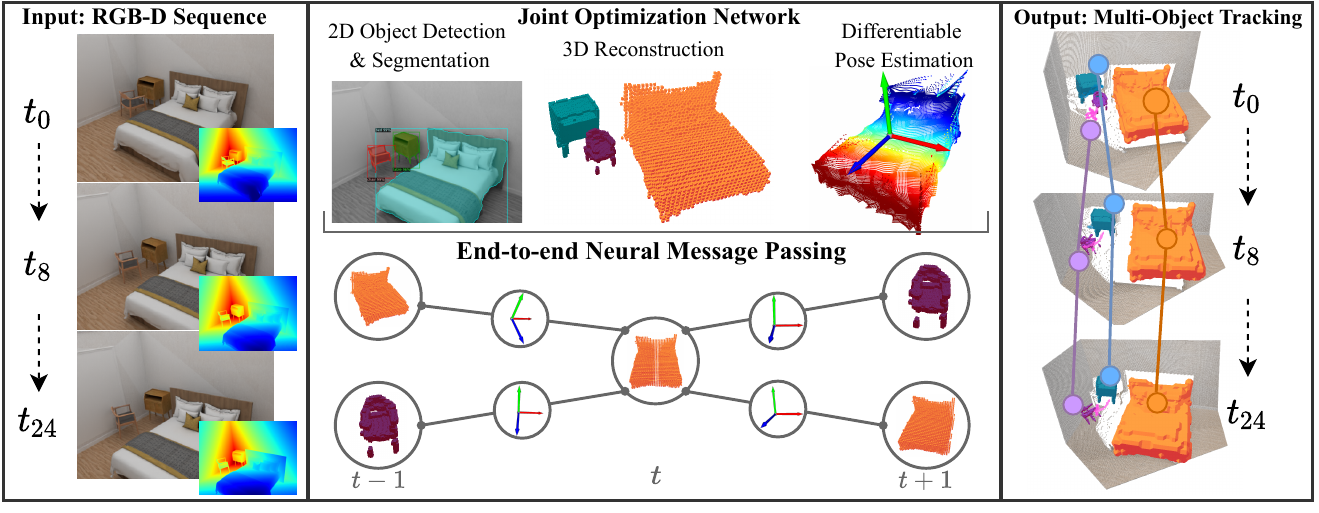}
  \caption{We investigate the task of 3D multi-object tracking using a novel synthetic indoor scene dataset. Our network leverages a 2D detection backbone with additional NOC~\cite{nocspaper} prediction and 3D reconstruction heads to predict per-object dense correspondences maps and 7-DoF pose parameters. We leverage those correspondences in our neural message passing based, fully end-to-end learnable network to model dependencies between objects over time for consistent multi-object tracking.}
  \label{fig:teaser}
\end{figure}

In order to train and evaluate MOT in the indoor setting, we introduce \DATASET, a new synthetic dataset consisting of 2381 unique indoor sequences with a total of 60k rendered RGB-D images together with corresponding instance semantics. For each sequence, we leverage scene layouts and 3D assets from 3D-FRONT to generate camera and object trajectories. 
Based on this data, we conducted a series of ablation studies which show that our holistic approach for differentiable pose estimation, 3D reconstruction and object association provides a significant improvement.  
Overall, our method improves the accumulated MOTA score by 24.8\% over existing state-of-the-art.  

In summary, our main contributions are as follows\footnote{Project page can be found at \url{https://domischmauser.github.io/3D_MOT/}}:
\begin{itemize}
    \vspace{-4px}
    \item A novel graph-based 3D tracking method with differentiable pose estimation for temporally-consistent object pose trajectories.
    \vspace{-4px}
    \item A new formulation for joint object completion and pose estimation over time by inter-frame message passing for improved data association.
    \vspace{-4px}
    \item A new synthetic dataset comprising extensive 2D and 3D annotations of indoor scenes with multiple moving objects and camera over time.
\end{itemize}

\section{Related Work\label{works}}

\subsection{RGB-D Object Tracking}
With the wide availability of consumer-grade RGB-D sensors, several approaches have been proposed to understand dynamic environments by object tracking. 
Many SLAM-based systems~\cite{tateno20162, runz2017co, salas2013slam++, mccormac2018fusion++} perform instance segmentation, pose estimation, to achieve object tracking. Dynamicfusion~\cite{newcombe2015dynamicfusion} introduces the reconstruction and tracking of dynamic, non-rigid scenes in SLAM by decomposing a non-rigidly deforming scene into a rigid canonical space and include moving objects. 
In addition, several approaches have been dedicated to perform MOT in dynamic scenes.
EM-Fusion~\cite{strecke2019fusion} proposes a probabilistic expectation maximization formulation (EM) to conduct object-level SLAM for data association.
Mask-Fusion~\cite{runz2018maskfusion} performs tracking based on optimizing the iterative closest point (ICP) error and photometric cost; however, Mask-Fusion relies on hand-crafted features for deciding non-static objects. 
MID-Fusion~\cite{xu2019mid}, introduced by Xu et al., uses an octree-based method to generate an object-level volumetric map and perform RGB-D camera tracking and object pose estimation with an ICP-algorithm.

In order to overcome reconstruction and object association issues, Müller et al.~\cite{muller2020seeing} propose a framework that jointly performs object completion and pose estimation, where objects are associated in a frame-to-frame fashion levering hand-crafted heuristics for object assignment. In contrast, our method learns optimal object associations over time in an end-to-end fashion, using a neural message passing network while performing fully differentiable pose estimation and 3D reconstruction. 

\subsection{Multi-object tracking with graphs and transformers}
A common approach for MOT is the tracking-by-detection paradigm~\cite{bewley2016simple, blmot, yin2021center}. Here,  objects are usually first localized in each frame by an object detector, followed by associations of proposals in adjacent frames to generate the tracking results. For this, Kalman Filters~\cite{WelchB95} or similarity measures~\cite{deepsort, tracktor_2019_ICCV}
association metric together with the Hungarian algorithm~\cite{kuhn1955} are leveraged for track association.

In recent years, several graph-based approaches have been introduced to perform data association. Bras\'o et al.~\cite{braso2020learning} propose a fully differentiable framework based on a message passing network modeling temporal dependencies to perform MOT for outdoor scenes. Yu et al.~\cite{yu2020spatio} additionally model spatial context with a second graph network for the spatial domain and incorporate a self-attention mechanism in both graphs for improved context learning. 
Novel transformer-based MOT frameworks perform multi-object tracking in a frame-to-frame fashion employing the concept of autoregressive track queries~\cite{zeng2021motr, meinhardt2021trackformer}.
These works perform tracking in a 2D space while ignoring the 3D spatial relations between objects and their 3D geometry. We instead utilize 3D pose and geometry features to model scene configurations over time.
For graph-based 3D tracking, Wang et al.~\cite{jointobjdetmot} propose a framework that jointly optimizes object detection and data association. Additionally, GNN3DMOT~\cite{gnn3dmot} introduces a graph neural network using 2D and 3D features for the MOT domain. They show that spatial and temporal interaction of the 2D and 3D object features can improve the tracking performance. 
Our method has similar motivation as prior graph-based MOT methods, but our differentiable optimization for object poses and geometry facilitates end-to-end learning, thus improving overall tracking performance.

\subsection{MOT datasets}
Recently, many new datasets have been proposed to facilitate research in the MOT domain, primarily for outdoor applications~\cite{voigtlaender2019mots, geiger2012we, milan2016mot16, caesar2020nuscenes, sun2020scalability, fabbri2021motsynth}. 
MOTS~\cite{voigtlaender2019mots} which is based on the outdoor dataset KITTI~\cite{geiger2012we}, is the first dataset including annotations for multi-object tracking and additional pixel-level instance mask annotations. The MOTChallenge~\cite{milan2016mot16} is a popular benchmark providing multiple datasets for 2D MOT with annotated pedestrians in crowded outdoor scenes. 
However, these datasets do not contain any 3D annotations, which makes tracking in the 3D domain infeasible. 
Other popular datasets such as KITTI~\cite{geiger2012we}, nuScenes~\cite{caesar2020nuscenes} and Waymo Open~\cite{sun2020scalability} contain 3D bounding box annotations, but lack pose and instance segmentation labels. The recently published MOTSynth dataset~\cite{fabbri2021motsynth} has diverse 3D annotations, yet does not provide the 3D geometry of objects. While existing datasets have been created for outdoor MOT, to the best of our knowledge, a publicly available dataset for indoor 3D MOT currently does not exist. Hence, we believe our new dataset \DATASET, providing complete 2D and 3D annotations will help to drive forward future research in the domain of indoor 3D MOT.
\section{\DATASET: Synthetic indoor MOT dataset\label{datasets}}

\begin{wraptable}{TL}{0.38\textwidth}
    \vspace{-12px}
    \begin{minipage}{0.38\textwidth}
      \caption{Dataset statistics of our synthetic indoor scene dataset \DATASET}
      \label{sample-table}
      \centering
        \begin{tabular}{lccc}
        \toprule
        Num of chairs & 4,210 \\
        Num of tables & 1,820 \\
        Num of sofas & 2,161 \\
        Num of beds & 449 \\
        Num of tv stands & 216 \\
        Num of wine coolers & 31 \\
        Num of nightstands & 140 \\
        \hline
        Dataset size (num scenes) & 2,381 \\
        Avg. num obj. per scene & 3.8 \\
        \bottomrule
        \end{tabular}
        \label{tab:data_statistics}
    \end{minipage}
    \vspace{-10px}
 \end{wraptable}

\paragraph{Dataset Overview}
We propose a dynamic indoor MOT dataset \DATASET\footnote{Dataset download at \url{https://domischmauser.github.io/3D_MOT/}} based on the 3D-FRONT dataset~\cite{fu20203d}.
3D-FRONT is a large-scale, comprehensive repository containing 18797 rooms diversely furnished by 3D furniture objects. We use Blenderproc~\cite{denninger2019blenderproc}, a procedural pipeline based on the open-source platform Blender, to generate photo-realistic renderings of the 3D-FRONT scenes. To the best of our knowledge, there is currently no publicly available dataset with extensive 2D and 3D annotations that depicts dynamically moving objects with a moving camera in indoor scenes.

\DATASET provides photo-realistic RGB-D images with their corresponding instance segmentation masks, class labels, 2D \& 3D bounding boxes, 3D geometry (voxel grids), 3D poses (NOCs maps) and camera parameters. 
Our dataset comprises $2,381$ sequences with a total of 60k images. Each sequence contains 25 frames and depicts different types of rooms with up to 5 moving objects belonging to 7 distinct object categories: chair, table, sofa, bed, TV stand, wine cooler, nightstand. 
The dataset statistics can be found in Table~\ref{tab:data_statistics}.
Our dataset comprises diverse object shapes with textured backgrounds, unlike data from~\cite{dahnert2021panoptic} which is restricted by white floors and walls.
We synthetically generate physically plausible camera and object motion to create the desired image sequence with realistically moving objects and moving camera over time.

\paragraph{Data Generation} 
Dynamic data is generated automatically, without any manual labeling or human supervision, and instead relying on sampling techniques and score-based evaluation metrics to guarantee the generation of plausible data sequences. Several approaches~\cite{chen2020learning, wang2019normalized} propose to map object instances to a Normalized Object Coordinate Space (NOCs) to infer object poses by predicting NOCs maps. Furthermore, we generate ground truth voxel grids in a $32^3$ resolution for 3D reconstruction.

\textbf{\begin{wrapfigure}{R}{0.55\textwidth}
    \vspace{-25px}
    \begin{minipage}{0.55\textwidth}
      \begin{algorithm}[H]
        \caption{MOT sampling}
        \label{algo:mot}
        \begin{algorithmic}[1]
        \Require $N_{\text{obj}} \geq 3$
        \State \text{Initialize} $\{ {}^0\bm{T}_{obj}^k\}_{k=1}^K $ \text{,} $ {}^0\bm{T}_{cam} $ \text{,} $ {}^0\epsilon $
        \For{$n \leftarrow 1$ \text{to} $N_{\text{max}}$}
            \State \text{Draw} $\bm{x}_{\text{obj}} \sim U([-\sigma, \sigma]^3)$
            \State \text{Draw} $\bm{\theta}_{\text{obj}} \sim U([-\phi_{\text{obj}}, \phi_{\text{obj}}]^3)$
            \If{ \text{collision free}} 
                \State ${}^{i+1}\bm{T}_{obj}^{k} \leftarrow {}^i\bm{T}_{obj}^{k} \cdot \bm{T}(\bm{x}_{\text{obj}},\bm{\theta}_{\text{obj}})$
                \For{$i \leftarrow 1$ \text{to} $N_{max}$}
                    \State ${}^{i}\epsilon \leftarrow {}^{0}\epsilon \cdot [1 + \text{log}(i + 1)] $
                    \State \text{Draw} $\bm{x}_{\text{cam}} \sim U([-{}^{i}\epsilon, {}^{i}\epsilon]^3)$
                    \State \text{Draw} $\bm{\theta}_{\text{cam}}  \sim U([-\phi_{\text{cam}}, \phi_{\text{cam}}]^3)$
                    \If{ \text{interest score}}
                        \State ${}^{i+1}\bm{T}_{cam} \leftarrow {}^{i}\bm{T}_{cam} \cdot \bm{T}(\bm{x}_{\text{cam}},\bm{\theta}_{\text{cam}})$
                    \EndIf
                \EndFor
            \EndIf
        \EndFor
        \end{algorithmic}
      \end{algorithm}
    \end{minipage}
    \vspace{-5px}
 \end{wrapfigure}}

We randomly choose rooms that contain at least 5 objects and generate one sequence per room. We randomly choose at least 3 objects ($N_{\text{obj}}$) and sample their location and orientation along with that of the camera 25 times to generate one tracking sequence. The detailed object and camera sampling can be seen in Algorithm~\ref{algo:mot}. The 3D position is denoted as $\bm{x} \in \mathbb{R}^3$ and the Euler orientation as $\bm{\theta} \in \mathbb{R}^3$. We first perform object position sampling: we weight the sampling direction with a factor $\sigma(n, d(\bm{x}, \bm{x'}))$ (see eq.~\ref{eq:repulse_weight}), similar to the repulsive potential~\cite{choset2005principles}, which is dependent on the distance from current object $\bm{x}$ to nearest obstacle $\bm{x}'$. This will guide the objects from each other and all obstacles away when within some predefined threshold $d^*$ to achieve larger motions. Additionally, we use Bezier interpolation to smooth the object trajectories. For camera pose sampling, we uniformly sample camera orientation angles between a predefined range $\phi$.
Camera position sampling is also weighted with a factor ${}^{i}\epsilon$, which gradually increases but is bounded to give the camera more freedom during the sampling. Admissible camera poses should surpass a minimum interest score threshold which is computed as the weighted sum of visible objects in that view. Views capturing moving objects receive a high interest score to guarantee meaningful renderings with enough objects. We restrict the number of maximal tries $N_{\text{max}}$ for camera and object sampling to 500.
\begin{equation} \label{eq:repulse_weight}
  \sigma(n, d(\bm{x}, \bm{x'})) =
    \begin{cases}
    \frac{1}{2} \sigma_0 \left( \frac{1}{d(\bm{x}, \bm{x}')} - \frac{1}{d^*} \right)^2 & \text{if $d(\bm{x}, \bm{x}')<d^*$ and $n<N_{max}$}\\
    1, & \text{else} 
    \end{cases}
\end{equation}

\begin{figure}[t]
  \centering
  \includegraphics[trim=0 3 32 3, clip, width=1\textwidth]{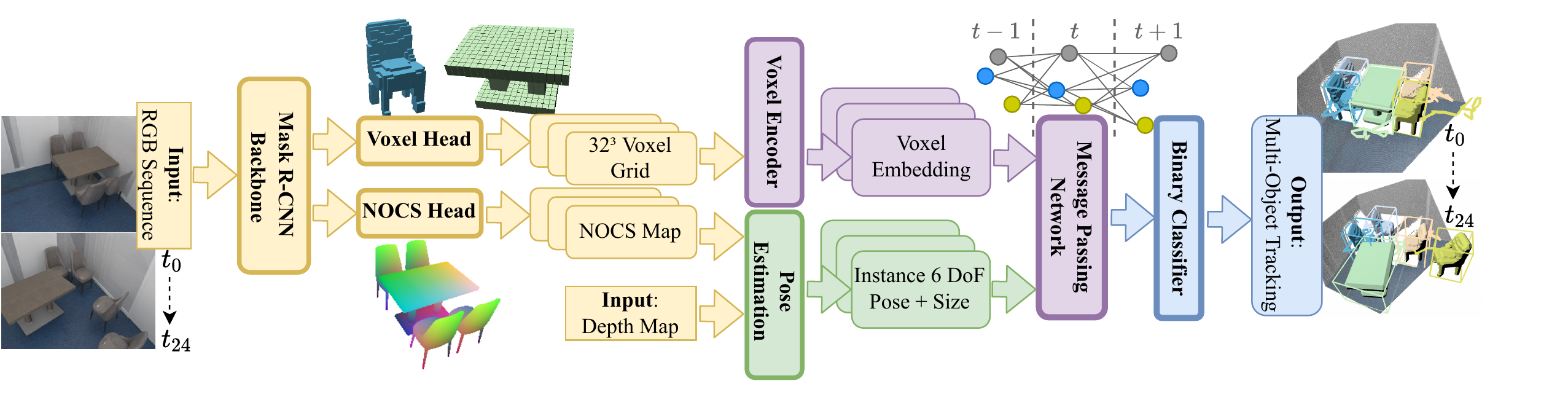}
  \caption{Network architecture overview: a Mask R-CNN backbone detects object instances from an RGB sequence, additionally predicting a $32^3$ voxel grid and a NOCs map for each object which is aligned by a pose estimation module using supplementary depth sequence data represented as a depth point cloud, outputting a 7-DoF pose. A 3D convolutional network encodes predicted voxel grids into initial node embeddings. For a consistent object tracking we employ a graph neural network with a consecutive binary classifier, predicting active and non-active graph connections, which is constructed from geometry and pose embeddings and jointly optimizes reconstruction, pose, and object association.}
  \label{fig:framework}
  \vspace{-0.3cm}
\end{figure}
\newpage
\section{Method Overview\label{method}}

From an input RGB-D sequence, our method predicts multi-object tracking of the objects observed in the sequence.
Our network architecture (Fig.~\ref{fig:framework}) consists of a Mask R-CNN backbone~\cite{he2017mask}, a 3D reconstruction network, a pose fitting pipeline and a neural message passing network for a subsequent multi-object tracking.
The network backbone takes as input the RGB sequence and performs 2D detection and instance segmentation for each image. Detected objects are then associated using input depth data by a neural message passing network in an end-to-end fashion, which enables joint optimization of object pose and 3D geometry for temporal consistent trajectories.

\subsection{2D Object Detection \& 3D Reconstruction}
We extend a Mask R-CNN backbone~\cite{he2017mask} with two additional heads: a \textit{voxel head} and a \textit{NOCs head}. The voxel head takes as input the predicted instance image patch $I_{\text{RGB}}$ from the box \& segmentation head and conducts both 3D object reconstruction and shape completion, outputting a $32^3$ per object voxel grid $O$. The NOCs head processes the instance image patch $I_{\text{RGB}}$ in parallel and outputs a Normalized Object Coordinate space (NOCs) patch $I_{\text{NOCs}} \in \mathbb{R}^{3 \times w_{\text{box}} \times h_{\text{box}}}$~\cite{nocspaper} with $w_{\text{box}}$ and $h_{\text{box}}$ as the bounding box dimensions, containing pose information of the predicted object.
The voxel and NOCs heads follow a decoder structure and take the same RoI (region of interest) feature embedding, computed by a RoI align operation, to predict voxel grid $O$ and NOCs patch $I_{\text{NOCs}}$, respectively. 
Our voxel head initially reshapes the RoI embedding $\bm{e}_{\text{RoI}}\in\mathbb{R}^{14\times14}$  into a $n_{\text{channels}}\times4^3$ embedding and reconstructs a $32^3$ voxel grid $O$ using a series of transposed 3D convolution layers with added 3D batch normalization. The NOCs head comprises multiple transposed 2D convolution layers with added 2D batch normalization, predicting from the RoI embedding $\bm{e}_{\text{RoI}}$ a $28\times28\times3$ NOCs map which is resized to the respective predicted bounding box size using a RoI align operation into a NOCs patch $I_{\text{NOCs}}$. 
During inference, we filter object detections utilizing non-maximum suppression between 2D bounding box predictions, as well as two thresholds $\kappa$ = 0.35 and $\nu$ = 0.35, discarding object detections with 2D bounding boxes which have a lower 2D IoU with any ground truth bounding box than $\kappa$ and objects with a lower objectness score than $\nu$.

\subsection{Differentiable Pose Estimation}
A pose estimation module utilizes the predicted NOCs patch $I_{\text{NOCs}}$, the depth map patch $I_{\text{Depth}}$ corresponding to the detected object, and camera intrinsics to infer frame-wise size $c^*$, location $\bm{t}^*$, and rotation $\bm{R}^*$ in a camera coordinate space for each object. Therefore, we backproject depth patch $I_{\text{Depth}}$ and NOCs patch $I_{\text{NOCs}}$ to point clouds $\bm{P_o}$ and $\bm{P_n}$. We perform statistical outlier removal on both point clouds, based on the distance to neighboring points, using a RANSAC outlier removal algorithm~\cite{Fischler1981RandomSC}.
This enables removing erroneous residuals which could potentially have a negative effect on the alignment. Finally, the Umeyama algorithm~\cite{umeyama1991} is employed to find the optimal 7-dimensional rigid transformation to align both cleaned point clouds:
\begin{equation}
c^*,\bm{R}^*,\bm{t}^* := \underset{c\in\mathbb{R}^+, \bm{R}\in SO_3, \bm{t}\in \mathbb{R}^3 }{\text{argmin}}{\|\bm{P_o} -  (c\bm{R}\cdot \bm{P_n} +\bm{t})\|}.
\end{equation}

\subsection{Neural Message Passing and Tracking}

We define a bidirectional graph neural network connecting object detections in consecutive frames within a window of 5 neighboring frames. A window size of 5 was selected since it enables a large receptive field for each graph node in the temporal domain to derive better features by having temporal context between frames. We initialize each graph node from the predicted object geometry and edge embeddings with encoded relative pose features. After a number of message passing steps $n_{\text{mp}}=4$, each edge embedding comprises information from neighboring nodes ensuring a temporal-context understanding. The final edge embeddings are classified by a binary classifier into active and non-active graph connections to predict unique tracklets. We assign ground truth instance ids to object detections by finding the maximum 3D IoU between a predicted bounding box with all possible ground truth bounding boxes in a frame and selecting its respective instance id. We discard graph connections between object detections which have a 3D IoU lower than a threshold $\tau$ = 0.05 for all ground truth bounding boxes. Object pairs with matching instance ids are assigned as positive (active) training pairs and objects pairs with distinct instance ids are assigned as negative (non-active) training pairs. 

\paragraph{Pose Embedding.} An edge of our message passing network consists of a relative pose embedding $\bm{e}_{ij} \in\mathbb{R}^{12}$ between two connected nodes $n_{i}$ and $n_{j}$.
The initial edge feature is computed by an MLP $\mathcal{N}_{\text{edge-enc}}$, encoding a concatenated feature vector with relative translation $\bm{t}=(x,y,z) \in \mathbb{R}^3$, relative rotation as euler representation $\bm{R}=(\alpha, \beta, \gamma) \in \mathbb{R}^3$, relative scale $c$ and relative time step $s$. 
\begin{equation}
\bm{e}_{ij} = \mathcal{N}_{\text{edge-enc}}([\bm{t}_{j} - \bm{t}_{i}, \bm{R}_{j} - \bm{R}_{i}, \log(\frac{c_{j}}{c_{i}}), s_{j} - s_{i}])
\end{equation}
\paragraph{Geometry Embedding.} A node feature of our message passing network consists of a shape embedding $\bm{a}_{i}\in\mathbb{R}^{16}$, encoded by a 3D convolutional network $\mathcal{N}_{\text{net-conv3D}}$ from the predicted $32^3$ object voxel grid $O$ outputted by the voxel head. The voxel encoder network employs a series of 3D convolutions, followed by a flatten operation with two consecutive affine layers and leaky ReLU as non-linearities. 

\subsection{Training and Inference}

We train our end-to-end approach on a single RTX A4000 with a batch size of 4. 
We first independently train our object detection, 3D reconstruction and pose estimation pipeline for 60 epochs with a learning rate of 0.008, Adam optimizer and L2-regularization of 0.0005 to ensure stable object detections with accurate geometry and pose predictions.
Additionally, we pre-train the tracking pipeline for 40 epochs with a learning rate of 0.001, Adam optimizer, and a L2-regularization of 0.001, using fixed pose and geometry features. Finally, our network is trained in an end-to-end fashion for 20 more epochs to jointly optimize object detections, 3D reconstructions, 7-DoF poses, and neural message passing to achieve the best performance. 

We guide the model to extract per-frame object information by the loss $\mathcal{L}_{\text{obj}}$ which we define as a weighted sum of the detection loss $\mathcal{L}_{\text{det}}$ proposed by Mask R-CNN~\cite{he2017mask}, a reconstruction loss $\mathcal{L}_{\text{rec}}$ and a NOCs loss $\mathcal{L}_{\text{noc}}$ for correspondence matching. 
The reconstruction loss $\mathcal{L}_{\text{rec}}$ is defined by a balanced binary cross-entropy loss (BCE), balancing occupied $O_{\text{occ}}$ and non-occupied voxels $O_{\text{free}}$ for a larger weighting of occupied areas utilizing an object dependent weighting $w_{\text{occ}}$. The NOCs loss $\mathcal{L}_{\text{noc}}$ is a symmetrical smooth-L1 loss which considers object symmetries for the predicted object class $pred_{cls}$ \textit{table} by choosing the minimal loss between ground truth NOCs patch $I_{\text{NOCs}}^{\text{gt}}$ and predicted NOCs patch $I_{\text{NOCs}}^i$ for all possible target rotations $i \in$ (0\degree, 180\degree).
\begin{align*} 
\mathcal{L}_{\text{noc}} &= \begin{cases}
            \text{min}_{i = [0\degree, 180\degree]}\hspace{0.1cm}L1_{\text{smooth}}(I_{\text{NOCs}}^i, I_{\text{NOCs}}^{\text{gt}})\hspace{0.2cm} \text{if $pred_{cls}$ == $table$}\\
            L1_{\text{smooth}}(I_{\text{NOCs}}, I_{\text{NOCs}}^{\text{gt}})\hspace{0.2cm} \text{else}
            \end{cases}
            \\
\mathcal{L}_{\text{rec}} &=  w_{\text{occ}} \cdot \text{BCE}(O_{\text{occ}}, O_{\text{occ}}^t) + \text{BCE}(O_{\text{free}}, O_{\text{free}}^t) \\
\mathcal{L}_{\text{obj}} &= \mathcal{L}_{\text{det}} +  3\cdot\mathcal{L}_{\text{noc}} + 0.75\cdot\mathcal{L}_{\text{rec}}
\end{align*}

For our tracking pipeline, we employ a binary cross-entropy loss with a weighting factor $w_{\text{act}}$ to account for the high imbalance between active graph connections $\bm{e}_{\text{act}}$ (GT associations) and inactive connections $\bm{e}_{\text{non-act}}$. For a final multi-object tracking across a sequence, we associate object detections by connecting active edges of the graph to trajectories. Nodes with no prior connections create a new trajectory if their instance id does not already exist in any other trajectory. In each time step, we extend tracklets according to the predicted associations. In case of non-unique assignment, we select the closest detection in terms of center distance.
\begin{align*}
    \mathcal{L}_{\text{track}} &=  w_{\text{act}} \cdot \text{BCE}(\bm{e}_{\text{act}}, \bm{e}_{\text{act}}^t) + \text{BCE}(\bm{e}_{\text{non-act}}, \bm{e}_{\text{non-act}}^t) 
\end{align*}
\paragraph{Evaluation metrics.} To evaluate our dynamic object
tracking, we adopt the Multiple Object Tracking Accuracy metric (MOTA)~\cite{bernardin2008evaluating}:
\begin{equation}
    \textrm{MOTA} = 1 - \frac{\sum_t(m_t + fp_t + mme_t)}{\sum_t gt_t}.
\end{equation}
where $m_t$, $fp_t$, $mme_t$, $gt_t$ are number of missed targets, false positives, identity switches and ground truth objects at time $t$. A match is considered positive if its L2 distance to the ground truth center is less than 40cm. We report the accumulated MOTA over all test sequences.
\section{Results\label{results}}
\begin{table*}[t]
\caption{Evaluation of MOTA, F1, Precision, and Recall on our  \DATASET dataset. We see that our end-to-end learnable approach outperforms a F2F-MaskRCNN baseline and current SOTA (SbO)~\cite{muller2020seeing}. Moreover, ablations show that joint optimization of the 7-DoF pose, 3D geometry and object associations over time via a message passing network improves tracking performance.}

\centering
\begin{tabular}{l|cccccc|c}
\hline
    & m$\downarrow$  & fp$\downarrow$ & mme $\downarrow$ & F1 $\uparrow$ & Precison $\uparrow$ & Recall $\uparrow$ & MOTA(\%)$\uparrow$  \\ \hline
F2F-MaskRCNN         & 13794          & 6107          & 645          & 0.721           & 0.795          & 0.662          & 46.2               \\
SbO~\cite{muller2020seeing}        & 12949          & 6400          & 802          & 0.724           & 0.777          & 0.677          & 46.7         \\
\hline\hline
Ours (no geometry)            & 10240          & 1749          & 58          & 0.824          & 0.928          & 0.747          & 68.5          \\
Ours (no joint opt.)            & 10025          & 1820          & 59          & 0.828          & 0.926          & 0.752          & 68.8          \\
Ours (no graph)   & 11068          & \textbf{1423}          & \textbf{47}          & 0.824          &  \textbf{0.940}          & 0.734          & 67.2          \\
Ours           & \textbf{8984} & 1873         & 58          & \textbf{0.841}       & 0.927  & \textbf{0.770} & \textbf{71.5} \\ \hline
\end{tabular}
\label{tab:mota_class}
\vspace{-0.3cm}
\end{table*}

\subsection{Quantitative Results\label{quant_res}} We compare our method against the current SOTA approach for indoor MOT, Seeing behind objects (SbO)~\cite{muller2020seeing}, which performs object detection directly in 3D and tracking via heuristic-based frame-wise matching. Additionally, we evaluate against a baseline F2F-MaskRCNN approach which conducts a frame-to-frame tracking based on point cloud matching by ICP with 2D detections from the same pretrained Mask R-CNN backbone as ours. Table~\ref{tab:mota_class} depicts quantitative results on our \DATASET test set of 398 sequences, evaluating F1, Precision, Recall and MOTA with the number of misses, false positives and mismatches. Our graph-based, end-to-end-learnable approach uses geometry and relative pose features between connected objects, achieving the best performance on all evaluation metrics, outperforming baselines in overall MOTA by 24.8\% and F1-score by 0.117. We refer to the supplemental material for additional class-specific tracking evaluations.

\subsection{Qualitative Results\label{qual_res}}
\begin{figure}[t]
  \centering
  \includegraphics[width=1\textwidth, trim=43 550 214 40, clip]{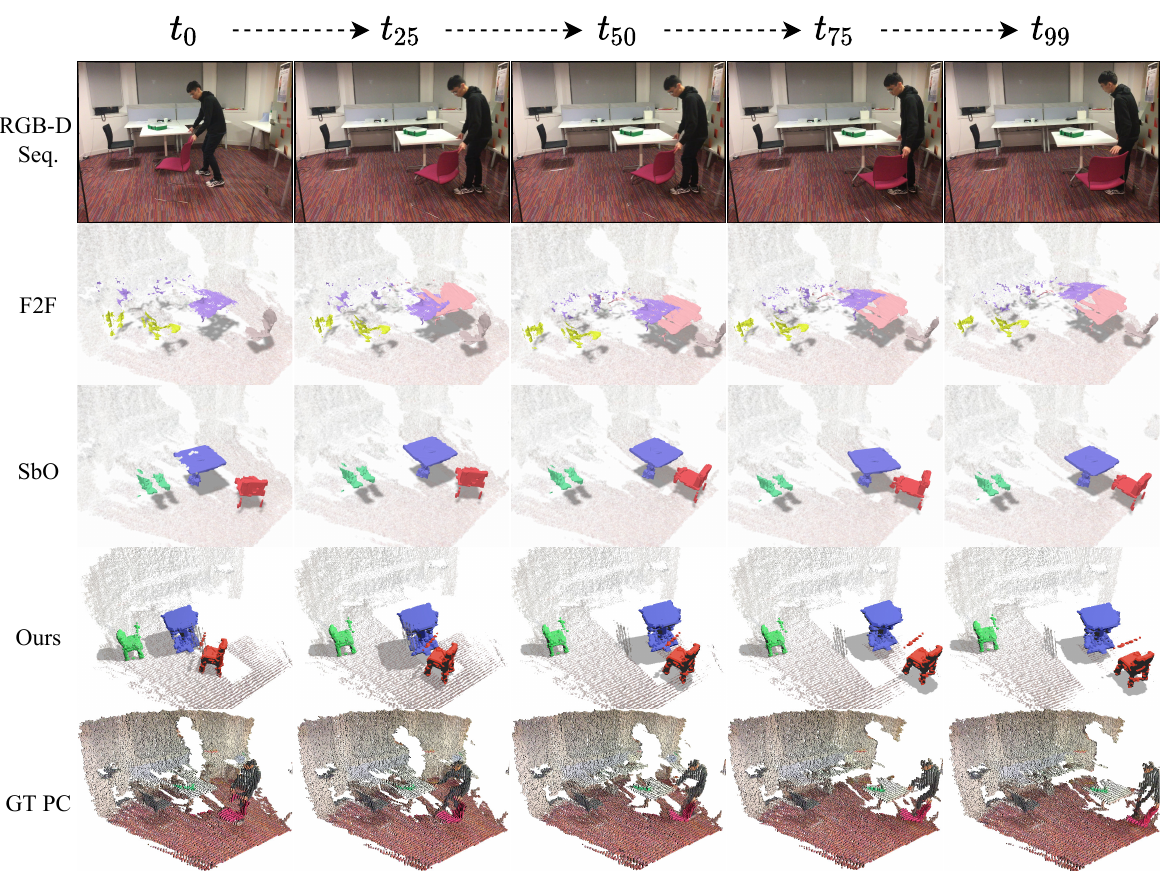}
  \caption{Qualitative comparison with SOTA method Seeing behind objects~\cite{muller2020seeing} and a F2F-MaskRCNN baseline for a scene from a real-world office dataset.}
  \label{fig:qual_res_real}
  \vspace{-0.5cm}
\end{figure}

\begin{figure}
\centering
\includegraphics[width=1\linewidth, trim=75 285 205 95, clip]{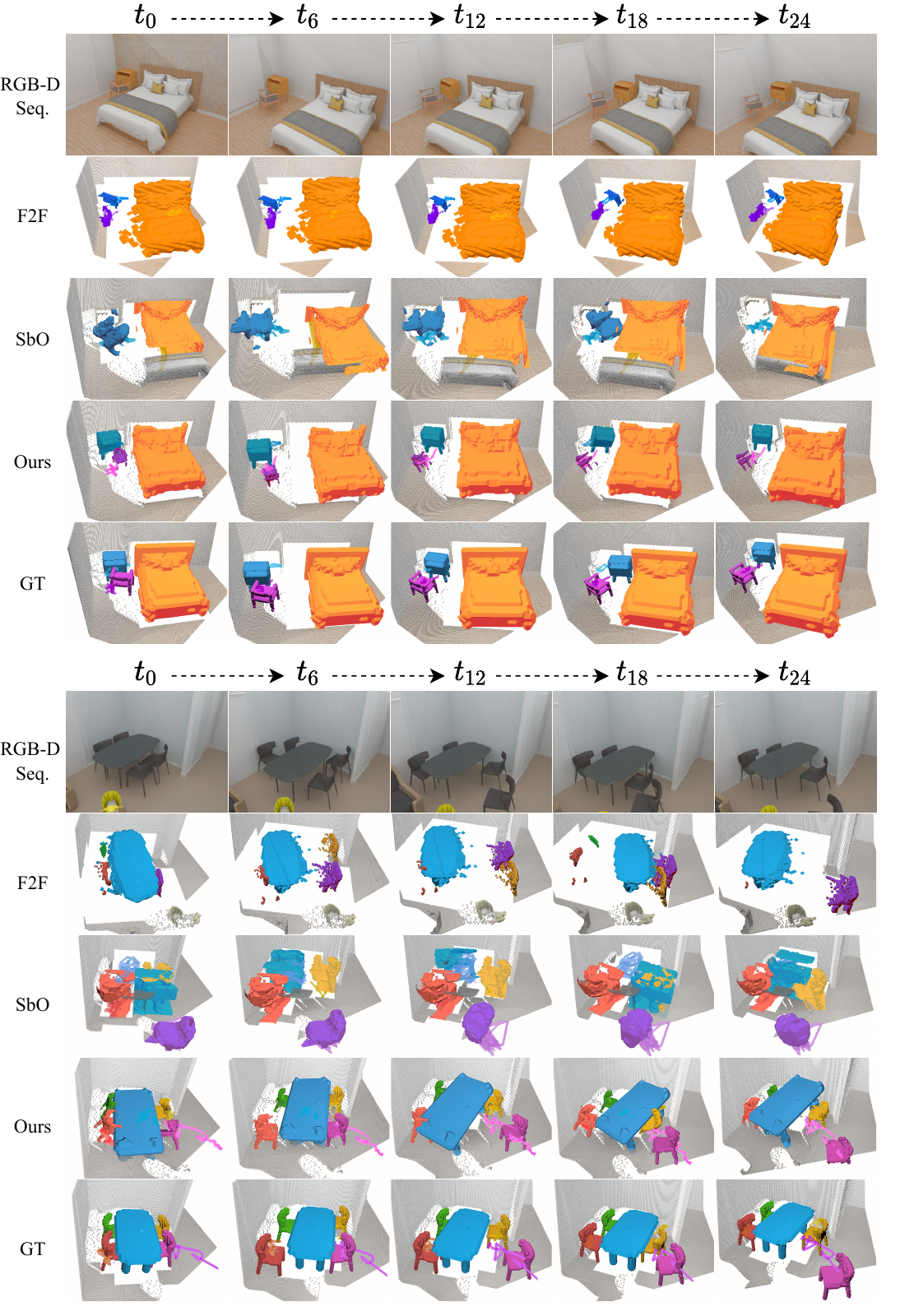}
\vspace{-0.3cm}
\caption{Qualitative comparison with Seeing behind objects (SbO)~\cite{muller2020seeing} and a F2F-MaskRCNN baseline on our synthetic \DATASET dataset. By jointly optimizing features over time, our approach predicts more consistent tracks for objects and achieves improved geometry completion and pose estimation accuracy. Object color encodings and line segments show instance id and estimated trajectories, respectively.
}
\label{fig:qual_res_syn}
\end{figure}

We further compare our approach qualitatively against the baselines on our synthetic indoor scene dataset \DATASET in Figure~\ref{fig:qual_res_syn} as well as on real-world office sequences from~\cite{muller2020seeing} in Figure~\ref{fig:qual_res_real}. 
Our approach is able to estimate accurate trajectories and shape reconstructions, even for heavily occluded objects such as the chairs in Figure~\ref{fig:qual_res_syn}. As the baselines are not optimized for temporal consistency, we observe more tracking failures in comparison to our method. 
Moreover, our approach yields higher reconstruction quality compared to \cite{muller2020seeing}, which does not optimize geometry over time. F2F Mask-RCNN reconstructs and tracks solely based on segmented object surfaces, often producing inaccurate geometry and pose estimates. 
Additionally, our method achieves more accurate reconstructions and more precise pose estimation over time on real-world sequences, as shown in Figure~\ref{fig:qual_res_real}.

\subsection{Ablations\label{ablations}} 
\paragraph{What is the effect of graph-based 3D tracking with differentiable pose estimation?}
When comparing our association step with a graph neural network (Ours) against an approach with a L2-distance heuristic (no graph), we obtain an increase in MOTA score by 4.3\% and a reduction of misses by 18.9\% (refer Table~\ref{tab:mota_class}). This confirms that our end-to-end trained, graph-based network can better learn consistent pose trajectories over time, enabling reasoning for trajectory matching considering an enlarged receptive field over multiple frames in contrast to a frame-by-frame heuristic.

\paragraph{Does end-to-end joint object completion and pose estimation over time improve tracking?}
We further analyze the effect of excluding geometry features in our tracking pipeline (no geometry). This results in a notable decrease in MOTA score by 3\% and recall by 2.3\% (refer Table~\ref{tab:mota_class}). We conclude that joint reconstruction and completion of the object geometry enables a more robust and geometrically more stable tracking. 
In particular, the effects of frequently occurring object occlusions can be alleviated, leading to a lower number of misses in trajectories. We further analyze the effect of training our pipeline end-to-end versus separate optimization (no joint opt.): Our jointly optimized approach improves MOTA by 2.7\% and achieves more reliable object detections and pose predictions. This shows that updating the object-level feature extraction steps based on the data association improves the final tracking.

\subsection{Limitations\label{limitations}}
While our approach presents a promising step to robust multi-object tracking, several limitations remain. As our reconstruction is limited by the dense voxel grid resolution, fine-scale details cannot be captured. Additionally, one could consider an appearance-based object representation that also models textures. This could further improve data association and lead to even more consistent object tracking by also optimizing for the appearance over time.

\subsection{Societal impact\label{soc_impact}}
This work proposes a method for multi-object tracking in indoor scenes. It can benefit XR applications and service robots to enable a better understanding of the dynamic environment. By joint reconstruction of the moving objects, it can enable 3D navigation and interaction with the tracked objects (like grasping, obstacle avoidance or digital replication of indoor scenes). For real-world applications, it requires careful consideration in terms of personal data privacy and potential bias towards certain object instances introduced by the training data.
\section{Conclusion\label{conclusion}}

We have introduced a new method for 3D multi-object tracking in RGB-D indoor scenes. By employing a graph-based,  end-to-end-learnable network with differentiable pose estimation and joint reconstruction, our method can predict robust object trajectories over time.
Experiments demonstrate a 24.8\% improvement in MOTA score over existing SOTA alternatives.
In a series of ablations, we conclude that by learning to optimize object poses and shapes over time, our method achieves temporally and spatially plausible trajectories. To train and evaluate our holistic approach, we introduce a novel synthetic MOT dataset \DATASET, with extensive 2D \& 3D annotations, which we hope will facilitate MOT research in the indoor setting. 
Overall, we believe our method is an important stepping stone for tracking and reconstruction of indoor environments.

\section{Acknowledgements}
\label{acknowledge}
This project is funded by the TUM Institute of Advanced Studies (TUM-IAS), the ERC Starting Grant Scan2CAD (804724), and the German Research Foundation (DFG) Grant Making Machine Learning on Static and Dynamic 3D Data Practical.

%%%%%%%%% REFERENCES
{\small
\bibliographystyle{plain}
\bibliography{refbib}
}
\newpage
\appendix

\section*{\Large\textbf{Appendix}}

In this appendix, we provide further details about our proposed method, additional quantitative and qualitative results and further information on our dataset \DATASET. 
\section{Details on \DATASET}
Our dataset \DATASET is created based on assets and scene layouts of the 3D-Front dataset~\cite{fu20203d}. The dataset is available at: \url{http://tiny.cc/MOTFront}.
\begin{figure}[h]
  \centering
  \includegraphics[width=1\textwidth]{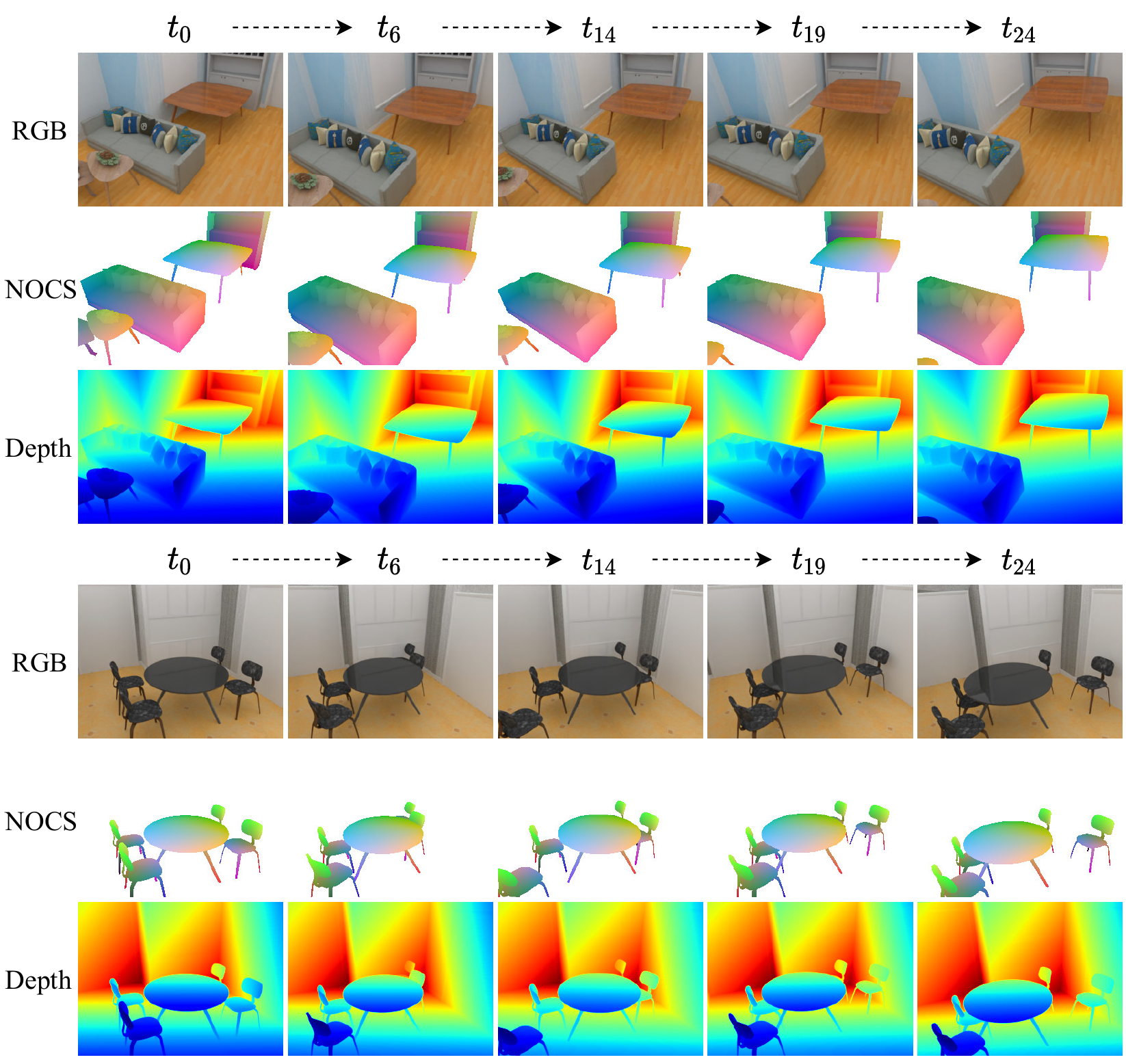}
  \caption{Additional sample sequences of our dataset \DATASET. The top row shows RGB images, the center row shows NOCs map images and the bottom row depicts the according depth map.}
  \label{fig:datasetseq_appendix}
\end{figure}

In Figure~\ref{fig:datasetseq_appendix}, two sample dataset sequences are depicted showing objects from different categories and distinct appearances. In general, our \DATASET dataset is very versatile, depicting different types of rooms (dining room, living room, bedroom) decorated with unique types of furniture. Our object and camera pose sampling algorithm generates dynamically moving objects and cameras. Some objects are heavily occluded, e.g. both chairs in the bottom dataset sequence, underlying the difficulty of 3D indoor tracking. For each scene, our dataset provides an \textit{annotation file} with 2D and 3D annotations, 25 \textit{RGB images}, 25 \textit{depth map renderings}, 25 \textit{NOCs map renderings} and the ground truth 3D geometry for each object.

\begin{itemize}
    \item \textit{Annotation file (per scene)}: Contains GT segmentation mask, class label, object location, rotation, scale, 2D \& 3D bounding boxes
    \vspace{-0.4cm}
    \item \textit{RGB images}: Photo-realistic renderings of an indoor scene
    \vspace{-0.4cm}
    \item \textit{Depth map}: Distance from the image plane
    \vspace{-0.4cm}
    \item \textit{NOCS images}: Object orientation
    \vspace{-0.4cm}
    \item \textit{Voxelized objects}: 3D object occupancy masks 
\end{itemize}

\section{Additional quantitative results}
Table~\ref{tab:iou_classwise} shows a class-wise comparison of 3D IoU scores between Seeing behind objects~\cite{muller2020seeing} and our approach. We compute the IoU score between the completed ground truth voxel grid and the predicted voxel grid for each object. By informing our shape feature extraction steps from the resulting data associations over time, our end-to-end learnable approach yields more fine-grained reconstructions in comparison to~\cite{muller2020seeing} which is reflected in higher IoU scores across all object classes and an increased overall 3D IoU score by 22.7\%. 

\begin{table*}[h]
\caption{Class-wise comparison of 3D IoU scores between Seeing behind objects~\cite{muller2020seeing} and our method, calculated from completed ground truth and predicted object voxel grid.}
\label{tab:iou_classwise} 
\centering
\resizebox{\textwidth}{!}{%
\begin{tabular}{l|ccccccc|c}
\hline
            3D IoU(\%)    & chair       & table         & sofa          & bed          & tv-stand          & cooler         & night-stand        & overall      \\ \hline
Seeing behind objects~\cite{muller2020seeing}       & 35.3          & 22.2      & 30.1         & 22.1          & 23.6           & 23.0          & 17.2          & 29.0         \\
Ours          & \textbf{53.3} & \textbf{39.2} & \textbf{60.1}          & \textbf{47.5}          & \textbf{39.8}       & \textbf{40.1}  & \textbf{29.0} & \textbf{51.7} \\ \hline
\end{tabular}}
\vspace{0.1cm}
\vspace{0.1cm}
\end{table*}

Furthermore, we depict overall and class-wise MOTA scores in comparison with our baselines in Table~\ref{tab:iou_classwise}. Our graph-based end-to-end learnable method achieves the highest MOTA scores for 3 out of the 7 object classes. Interestingly, our method performs best in the classes chair and table, the two largest classes with the most occurrences and the highest number of different shapes. Especially, for the most challenging object class chair which often occurs in scenes with multiple chairs located very close to each other, e.g. as shown in Figure~\ref{fig:datasetseq_appendix}, our method has an increased MOTA score by 19.6\% in comparison to Seeing behind objects~\cite{muller2020seeing}.

\begin{table*} [h]
\caption{Classwise MOTA score evaluation on our dataset \DATASET.}
\label{tab:mota_classwise} 
\centering
\resizebox{\textwidth}{!}{%
\begin{tabular}{l|ccccccc|c}
\hline
MOTA(\%)$\uparrow$    & chair       & table         & sofa          & bed          & tv-stand          & cooler         & night stand        & overall      \\ \hline
F2F-MaskRCNN        & 37.7          & 34.9          & 41.3         & 63.3          & 53.6           & 48.9          & 62.6          & 46.2               \\
Seeing behind objects~\cite{muller2020seeing}       & 56.8          & 33.3      & 39.5         & 18.9          & 52.1           & 42.2          & 36.2          & 46.7         \\
\hline\hline
Ours (no pose)          & 57.6          & 52.6          & 55.3          & 72.7          & 54.3          & 50.6          & 28.8          & 58.3          \\
Ours (no geometry)          & 71.8          & 58.3          & 59.3          & 77.7          & 56.9          & \textbf{52.3}          & 49.8          & 68.5          \\
Ours (no joint opt.)          & 72.4          &  \textbf{58.4}          & 59.3          & \textbf{77.8}          & 57.3          & \textbf{52.3}          & \textbf{50.1}          & 68.8          \\
Ours (no graph)          & 68.4          & 56.4          & \textbf{63.8}          & 73.2          & 56.9          & 47.7         & 47.4          & 67.2          \\
Ours          & \textbf{75.4} & \textbf{58.4} & 62.1          & 74.3          & \textbf{61.1}       & 39.2  & 50.0 & \textbf{71.5} \\ \hline
\end{tabular}}
\vspace{0.1cm}
\vspace{0.1cm}
\end{table*}

\section{Additional qualitative results}
\begin{figure}[H]
  \centering
  \includegraphics[width=1\textwidth, trim=10 0 0 0, clip]{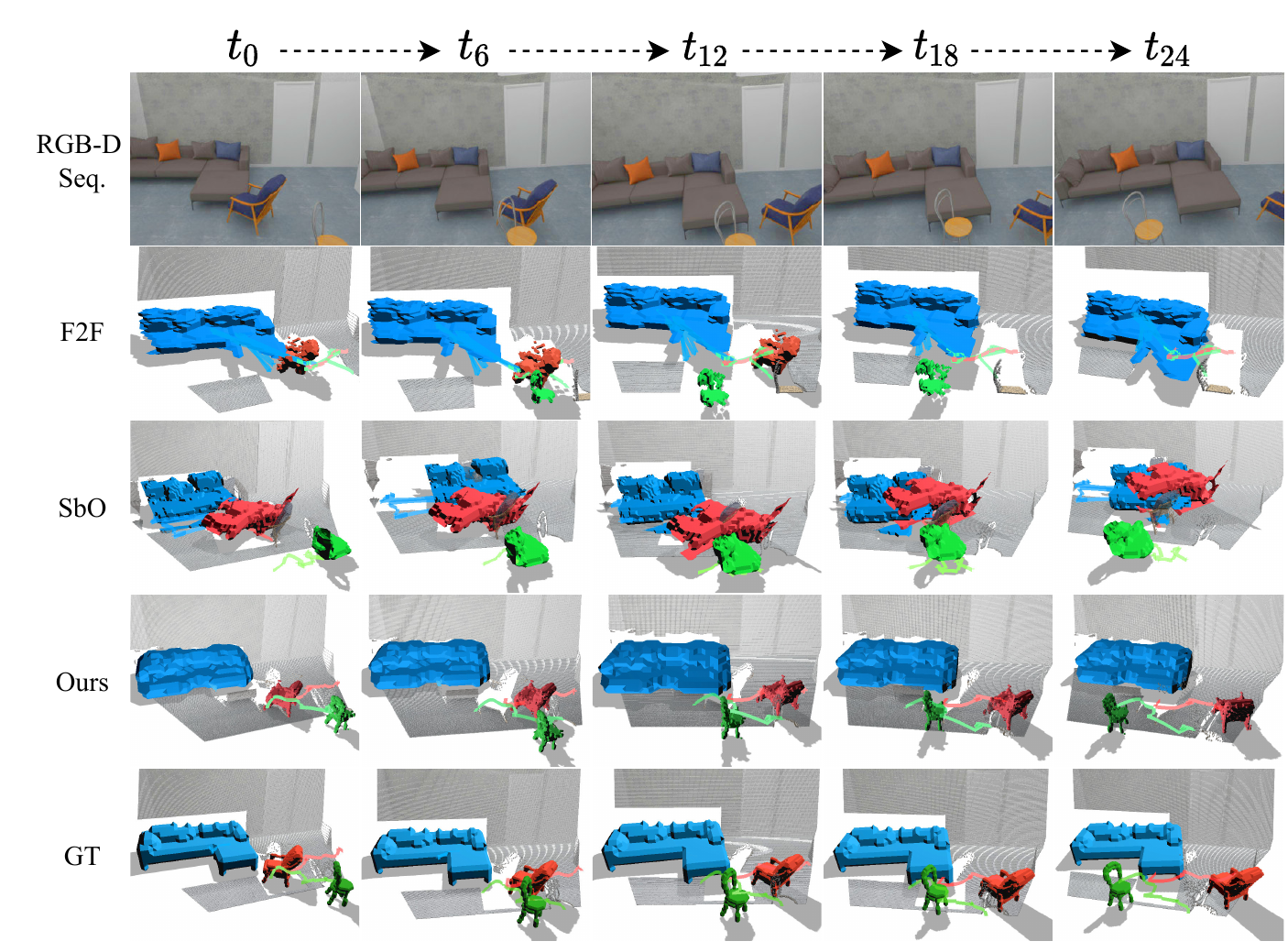}
  \caption{Further qualitative comparison with Seeing behind objects (SbO)~\cite{muller2020seeing} and a F2F-MaskRCNN baseline on our synthetic \DATASET dataset. Object color encodings and line segments show instance id and estimated trajectories, respectively.} %
  \label{fig:res_appendix}
\end{figure}

In Figure~\ref{fig:res_appendix}, we further show qualitative results depicting a living room scene with 3 moving objects. Our approach is able to predict detailed object shapes and robust pose trajectories for the whole sequence while F2F-MaskRCNN and Seeing behind objects have inconsistent object poses, missing detections and inaccurate 3D reconstructions.

\section{Implementation details}

\subsection{Network Details}
\begin{figure}[t]
  \centering
  \includegraphics[width=1\textwidth]{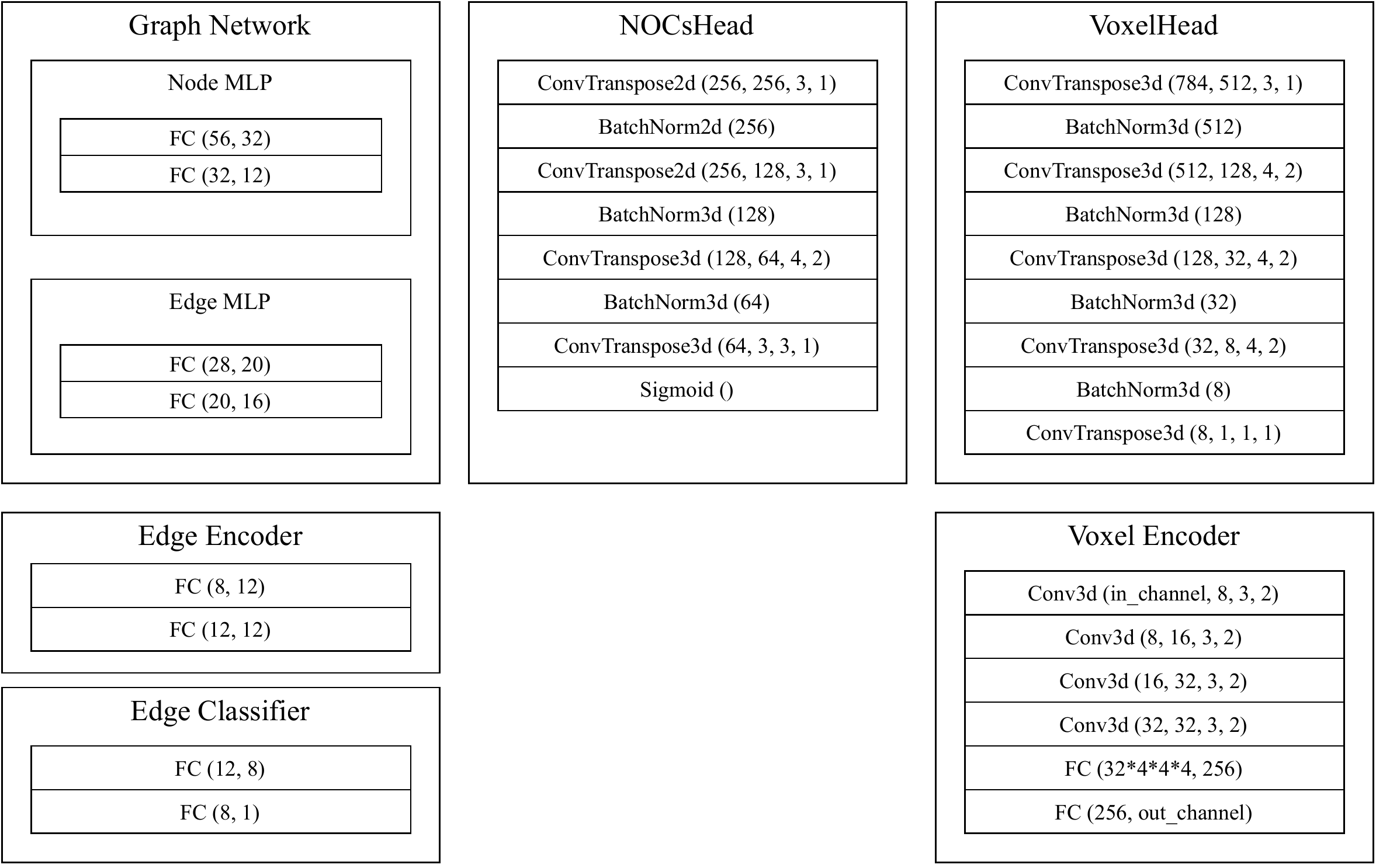}
  \caption{Network architecture specification.}
  \label{fig:net_appendix}
\end{figure}

In Figure~\ref{fig:net_appendix}, we detail the configuration of each network component. Besides the default Mask-RCNN backbone, we employ two more network heads NOCsHead and VoxelHead, and 4 additional networks: Voxel Encoder, Graph Network, Edge Encoder, and Edge Classifier. We provide the layer parameters as (input channels, output channels, kernel size, stride) and (input features, output features). 

\subsection{Pose estimation}
We employ two outlier removal steps to clean the predicted NOCs point cloud $\bm{P_n}$ and back-projected depth point cloud $\bm{P_o}$: first, a statistical outlier removal algorithm considers for each point a number of neighboring points $n_{\text{nbr}} = 20$, calculates the average distance for a given point and removes all points which are exceeding a standard deviation threshold $std_{\text{nbr}} = 2$. Second, the RANSAC outlier removal algorithm~\cite{Fischler1981RandomSC} selects an optimal set of corresponding points from the source and target point cloud derived from a minimal distance score utilizing linear least squares regression.

As a final step in our pose estimation pipeline, we transform the differentially optimized transformation matrix ${}^{\text{cam}}\bm{T}_{\text{pose}}$ $\in(c^*, \bm{t}^*, \bm{R}^*)$ from camera frame by multiplication with camera extrinsics into a uniform world coordinate space ${}^{\text{world}}\bm{T}_{\text{pose}}$ to ensure comparability between input frames of a sequence.

\subsection{Neural Message Passing}
Information of connected nodes is propagated by a series of message passing steps which is divided into two updates. First, a node to edge update is performed by a MLP taking as input the previous edge embedding and the two attached nodes. At a second step an edge to node update is conducted by a second MLP. Each MLP input is defined by a message which aggregates all neighboring edge embeddings by the permutation invariant \textit{mean} operation and the previous node embedding. After a number of message passing steps $n_{\text{mp}}=4$ each edge embedding comprises information from neighboring nodes ensuring a temporal-context understanding. 

\section{Code and Data}
Our dataset \DATASET is created based on the 3D-Front dataset~\cite{fu20203d}. We utilize code from the official implementations from Detectron2~\cite{wu2019detectron2} to train the Mask R-CNN backbone~\cite{he2017mask} and BlenderProc~\cite{denninger2019blenderproc} to render images.

\section{Licenses}
The 3D-Front dataset is distributed under the CC BY-NC-SA 4.0 license, we release \DATASET under the same licence and note that all rights remain with the owners of 3D-Front.
DLR-RM/BlenderProc is licensed under the GNU General Public license v3.0.
Detectron2 is released under the Apache 2.0 license.

\end{document}